\documentclass{article}
\usepackage[preprint]{colm2026_conference}

\usepackage{microtype}
\usepackage{hyperref}
\usepackage{url}
\usepackage{booktabs}
\usepackage{multirow}
\usepackage{graphicx}
\usepackage{float}
\usepackage{amsmath}
\usepackage{amssymb}

\definecolor{darkblue}{rgb}{0, 0, 0.5}
\hypersetup{colorlinks=true, citecolor=darkblue, linkcolor=darkblue, urlcolor=darkblue}

\usepackage{lineno}

\title{Fragile Reasoning: A Mechanistic Analysis of LLM Sensitivity to Meaning-Preserving Perturbations}

\author{
  \textbf{Shou-Tzu Han, Rodrigue Rizk, KC Santosh} \\
  Department of Computer Science \\
  University of South Dakota \\
  \texttt{shoutzu.han@usd.edu, rodrigue.rizk@usd.edu, kc.santosh@usd.edu}
}

\begin{document}

\ifcolmsubmission
\linenumbers
\fi

\maketitle

\begin{abstract}
Large language models demonstrate strong performance on mathematical reasoning benchmarks, yet remain surprisingly fragile to meaning-preserving surface perturbations. We systematically evaluate three open-weight LLMs, Mistral-7B \citep{jiang2023mistral7b}, Llama-3-8B \citep{grattafiori2024llama3herdmodels}, and Qwen2.5-7B \citep{qwen2025qwen25technicalreport}, on 677 GSM8K problems paired with semantically equivalent variants generated through name substitution and number format paraphrasing. All three models exhibit substantial answer-flip rates (28.8\%--45.1\%), with number paraphrasing consistently more disruptive than name swaps. To trace the mechanistic basis of these failures, we introduce the Mechanistic Perturbation Diagnostics (MPD) framework, combining logit lens analysis, activation patching, component ablation, and the Cascading Amplification Index (CAI) into a unified diagnostic pipeline. CAI, a novel metric quantifying layer-wise divergence amplification, outperforms first divergence layer as a failure predictor for two of three architectures (AUC up to 0.679). Logit lens reveals that flipped samples diverge from correct predictions at significantly earlier layers than stable samples. Activation patching reveals a stark architectural divide in failure \textit{localizability}: Llama-3 failures are recoverable by patching at specific layers (43/60 samples), while Mistral and Qwen failures are broadly distributed (3/60 and 0/60). Based on these diagnostic signals, we propose a mechanistic failure taxonomy (localized, distributed, and entangled) and validate it through targeted \textit{repair} experiments: steering vectors and layer fine-tuning recover 12.2\% of localized failures (Llama-3) but only 7.2\% of entangled (Qwen) and 5.2\% of distributed (Mistral) failures.
\end{abstract}

\section{Introduction}

Large language models (LLMs) have achieved impressive performance on mathematical reasoning benchmarks such as GSM8K \citep{cobbe2021trainingverifierssolvemath}, with recent open-weight models routinely exceeding 75\% accuracy. These results are often interpreted as evidence of emergent reasoning capabilities. However, a growing body of work questions whether such performance reflects genuine mathematical understanding or brittle pattern matching over surface-level features \citep{stolfo2023causalframeworkquantifyrobustness, zhang2024carefulexaminationlargelanguage}.

A simple diagnostic for robust reasoning is \textit{meaning-preserving perturbation}: if a model truly understands a math problem, substituting character names or rephrasing ``\$5'' as ``5 dollars'' should not alter the answer. Yet we find that such trivial modifications cause three widely-used LLMs to flip their answers on 28.8\%--45.1\% of problems, revealing a striking gap between benchmark accuracy and reasoning reliability.

Understanding \textit{why} these failures occur is as important as documenting them. Prior work has primarily characterized perturbation sensitivity at the behavioral level, measuring accuracy drops under various transformations \citep{shi2023largelanguagemodelseasily, zhou2023mathattackattackinglargelanguage}. We go further by applying mechanistic interpretability tools to trace where and how failures originate within model internals. Specifically, we combine three complementary techniques: \textbf{logit lens} analysis \citep{nostalgebraist2020logitlens} to identify the layer at which correct and incorrect predictions first diverge; \textbf{activation patching} \citep{meng2023locatingeditingfactualassociations} to test whether failures can be localized to specific token positions and layers; and \textbf{component ablation} to disentangle the causal contributions of attention heads and MLP sublayers.

Our analysis reveals that reasoning fragility is not a monolithic phenomenon. Across the three models, we find qualitatively different mechanistic failure signatures: Mistral-7B \citep{jiang2023mistral7b} exhibits the highest flip rate (45.1\%) with strongly distributed failures (3/60 patching recovery). Llama-3-8B \citep{grattafiori2024llama3herdmodels} shows a moderate flip rate (33.5\%) but highly localized failures (43/60 recovery). Qwen2.5-7B \citep{qwen2025qwen25technicalreport} has the lowest flip rate (28.8\%) with a distinctive MLP-dominant failure pattern (0/60 recovery). Across all three models, flipped samples diverge at significantly earlier layers ($p < 0.05$), and number format paraphrasing is consistently more disruptive than name substitution.

Our contributions are as follows:
\begin{enumerate}
    \item We propose a \textit{mechanistic failure taxonomy} (localized, distributed, entangled) with formal classification criteria, providing a principled framework for diagnosing perturbation-induced reasoning failures. We validate that this taxonomy predicts repair difficulty through targeted intervention experiments.
    \item We conduct a comparative cross-model mechanistic analysis of perturbation-induced reasoning failures, revealing that models with similar parameter counts and training paradigms exhibit qualitatively different internal failure signatures.
    \item We introduce the \textit{Cascading Amplification Index} (CAI), a quantitative diagnostic within the MPD framework that significantly predicts failure across all three architectures ($p < 0.05$), outperforming first divergence layer for models with distributed and localized failures (AUC up to 0.679).
    \item We establish the \textit{Mechanistic Perturbation Diagnostics} (MPD) framework, combining logit lens, activation patching, component ablation, and CAI into a reusable diagnostic pipeline applicable to any transformer-based model.
\end{enumerate}

\section{Related Work}

\paragraph{Robustness of LLM Reasoning.}
Prior work has documented LLM sensitivity to surface-level input variations \citep{ribeiro2020accuracybehavioraltestingnlp, wang2022adversarialgluemultitaskbenchmark}, including in mathematical reasoning where models rely on shallow heuristics \citep{stolfo2023causalframeworkquantifyrobustness}, are distracted by irrelevant context \citep{shi2023largelanguagemodelseasily}, fail under adversarial rephrasing \citep{zhou2023mathattackattackinglargelanguage}, may reflect memorization rather than generalization \citep{zhang2024carefulexaminationlargelanguage}, and degrade with increased input length \citep{levy2024tasktokensimpactinput}. These studies operate at the behavioral level; we extend them by tracing \textit{where} and \textit{how} failures arise mechanistically.

\paragraph{Mechanistic Interpretability.}
The logit lens \citep{nostalgebraist2020logitlens} and tuned lens \citep{belrose2025elicitinglatentpredictionstransformers} reveal how predictions evolve across layers. Activation patching \citep{meng2023locatingeditingfactualassociations, goldowskydill2023localizingmodelbehaviorpath} identifies causal subcomponents, while circuit discovery \citep{conmy2023automatedcircuitdiscoverymechanistic, wang2022interpretabilitywildcircuitindirect} and component ablation \citep{olsson2022incontextlearninginductionheads, geva2023dissectingrecallfactualassociations} disentangle architectural roles. We unify these into a diagnostic framework for perturbation-induced reasoning failure.

\paragraph{Mathematical Reasoning in LLMs.}
GSM8K \citep{cobbe2021trainingverifierssolvemath} is a standard reasoning benchmark where chain-of-thought prompting \citep{wei2023chainofthoughtpromptingelicitsreasoning, wang2023selfconsistencyimproveschainthought, kojima2023largelanguagemodelszeroshot} has driven rapid progress. However, recent work questions the depth of this reasoning \citep{wu2024reasoningrecitingexploringcapabilities, dziri2024faith}. Our perturbation framework offers a complementary test: if reasoning is robust, meaning-preserving changes should not alter answers.

\begin{table*}[t]
  \centering
  \small
  \setlength{\tabcolsep}{4pt}
  \begin{tabular}{lcccc}
    \toprule
    Work & Behavioral Eval & Mechanistic Diagnosis & Failure Taxonomy & Repair Validation \\
    \midrule
    Stolfo et al.\ (2023) & Yes & No  & No  & No \\
    Shi et al.\ (2023)    & Yes & No  & No  & No \\
    Zhou et al.\ (2023)   & Yes & No  & No  & No \\
    Levy et al.\ (2024)   & Yes & No  & No  & No \\
    Wu et al.\ (2024)     & Yes & No  & No  & No \\
    \midrule
    Ours                  & Yes & Yes & Yes & Yes \\
    \bottomrule
  \end{tabular}
  \caption{Comparison with prior work on LLM reasoning robustness. Prior studies mainly evaluate behavioral sensitivity under perturbations, whereas our work adds mechanistic diagnosis, failure taxonomy, and repair validation.}
  \label{tab:related_comparison}
\end{table*}

\section{Methodology}

Our methodology consists of three stages: (1) constructing a controlled perturbation dataset from GSM8K, (2) evaluating model robustness via paired inference, and (3) applying mechanistic interpretability techniques to diagnose the internal causes of failure.

\begin{figure*}[t]
  \centering
  \includegraphics[width=\textwidth,trim={20 20 40 20},clip]{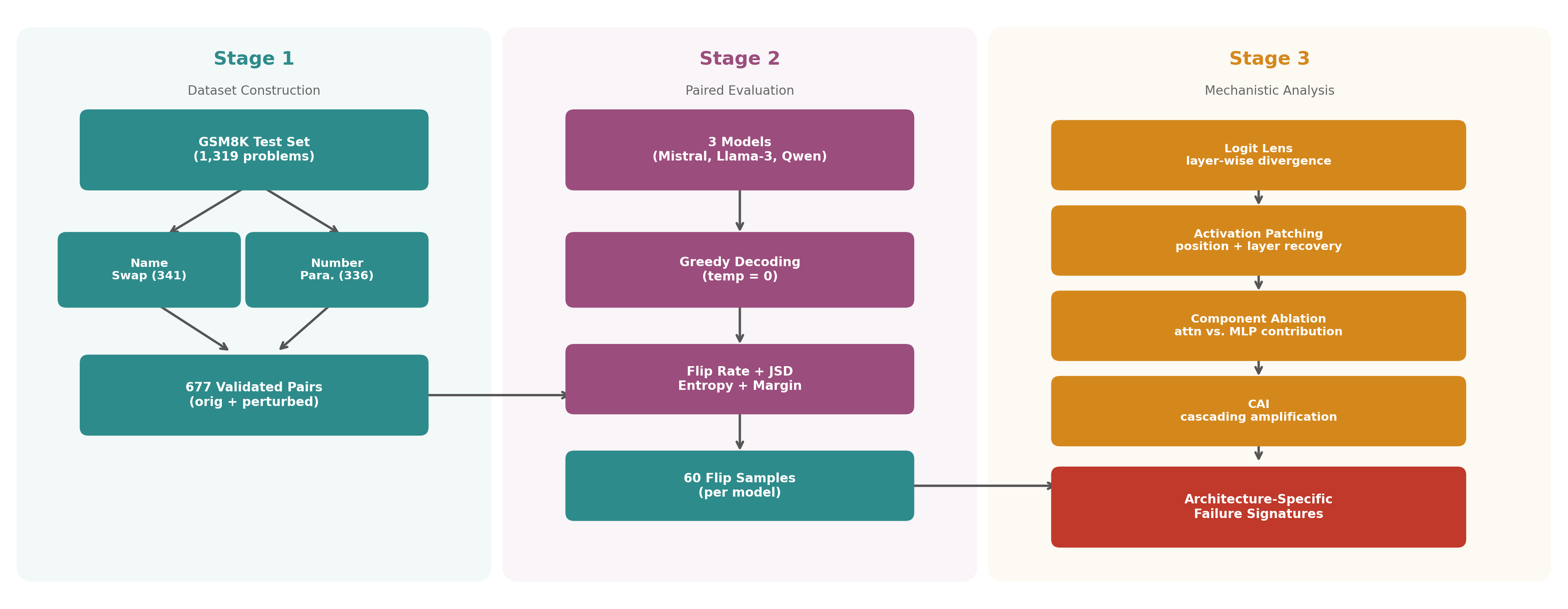}
  \caption{Overview of MPD. We construct meaning-preserving perturbation pairs from GSM8K, evaluate paired robustness across models, and diagnose failures with logit lens, activation patching, component ablation, and the Cascading Amplification Index (CAI).}
  \label{fig:pipeline}
\end{figure*}

\subsection{Perturbation Dataset Construction}

We construct a dataset of 677 valid problem pairs from the GSM8K test set \citep{cobbe2021trainingverifierssolvemath}, each consisting of an original problem and a meaning-preserving variant. We apply two perturbation types: \textbf{name substitution} (341 pairs), where all character names are replaced with randomly sampled alternatives, and \textbf{number format paraphrasing} (336 pairs), where numerical expressions are rephrased into equivalent formats (e.g., ``\textit{\$5}'' $\rightarrow$ ``\textit{5 dollars}'', ``\textit{60\%}'' $\rightarrow$ ``\textit{60 percent}''). Both types preserve the mathematical structure and correct answer. We discard any pair where the perturbation introduces ambiguity.

\subsection{Paired Robustness Evaluation}

For each model, we perform greedy decoding (temperature = 0) on both the original and perturbed problem using a fixed prompt instructing the model to solve the problem and output the answer after a \texttt{\#\#\#\#} delimiter. A sample $(q, q')$ is a \textit{flip} if the model answers the original correctly but the perturbed variant incorrectly:
\begin{equation}
    \text{flip}(q, q') = \mathbb{1}[f(q) = a^*] \cdot \mathbb{1}[f(q') \neq a^*]
\end{equation}

\subsection{Mechanistic Analysis}

We apply three complementary interpretability techniques to the subset of flipped samples.

\subsubsection{Logit Lens Analysis}

The logit lens \citep{nostalgebraist2020logitlens} projects the hidden state at each layer $l$ through the unembedding matrix $W_U$:
\begin{equation}
    p^{(l)} = \text{softmax}(W_U \cdot h^{(l)})
\end{equation}
We identify the \textit{first divergence layer}: the earliest layer at which the top-1 predicted token differs between the original and perturbed inputs, and compare distributions using a Mann-Whitney $U$ test.

\begin{figure}[t]
  \centering
  \includegraphics[width=\columnwidth]{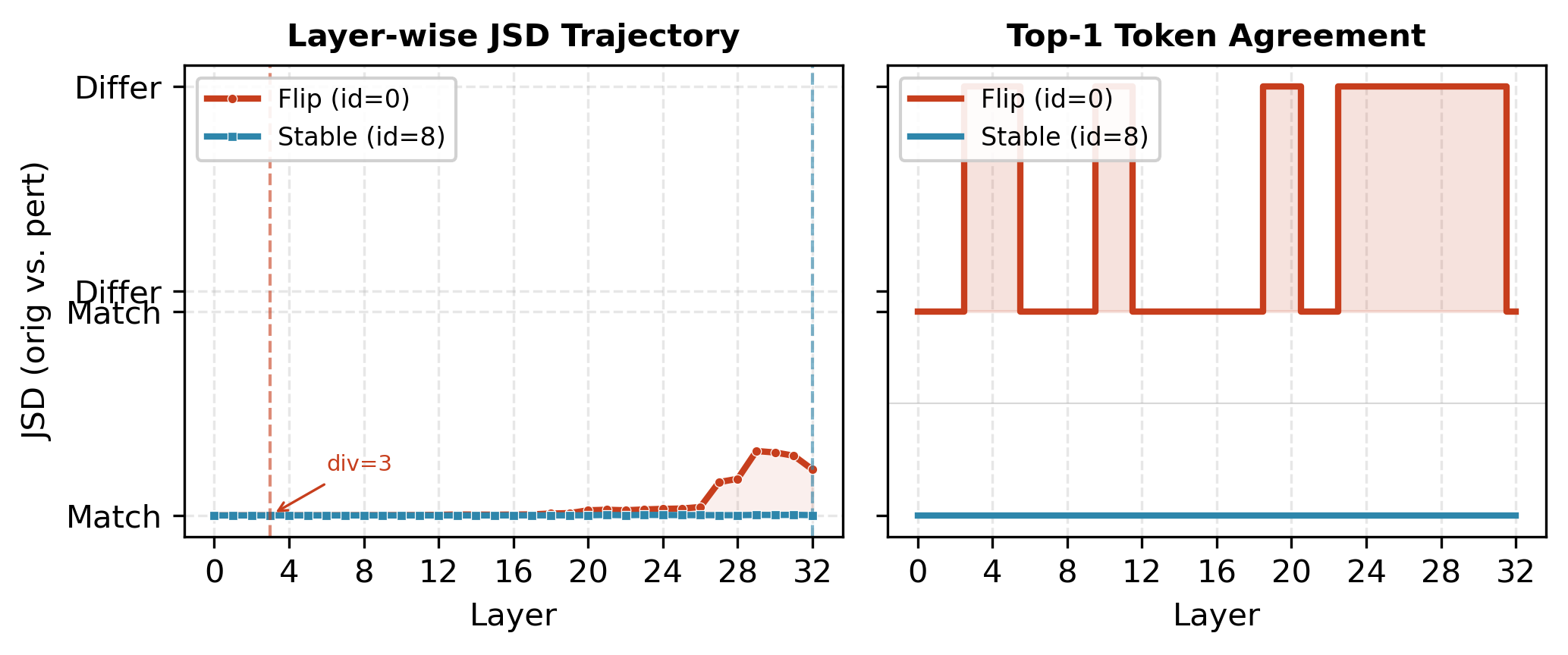}
  \caption{Logit lens applied to a flipped sample and a stable sample from Mistral-7B \citep{jiang2023mistral7b}. JSD rises sharply for the flip case but remains near zero for the stable case.}
  \label{fig:logit_lens_example}
\end{figure}

\subsubsection{Activation Patching}

For each flipped sample and each layer $l$, we replace the hidden states at diverging token positions with those from the original (correct) forward pass:
\begin{equation}
    \tilde{h}^{(l)}_t = \begin{cases}
        h^{(l)}_{\text{orig}, t} & \text{if } t \in \mathcal{T}_{\text{div}} \\
        h^{(l)}_{\text{pert}, t} & \text{otherwise}
    \end{cases}
\end{equation}
A high recovery rate indicates localized failures; a low rate indicates distributed sensitivity.

\subsubsection{Component Ablation}

For each layer $l$, we zero out the output of either the attention sublayer or MLP sublayer:
\begin{equation}
    h^{(l)} = h^{(l-1)} + \text{Attn}^{(l)}(h^{(l-1)}) + \text{MLP}^{(l)}(h^{(l-1)})
\end{equation}
If ablating component $c$ at layer $l$ yields the correct answer, then $c^{(l)}$ was causally contributing to the failure.

\subsubsection{Cascading Amplification Index}
\label{sec:cai}

To quantify whether layer-wise divergence amplifies or attenuates across the network, we introduce the \textit{Cascading Amplification Index} (CAI):
\begin{equation}
    \text{CAI} = \frac{1}{L - l_0} \sum_{l=l_0}^{L-1} \mathbb{1}[\Delta_{l+1} > \Delta_l]
\end{equation}
where $\Delta_l = \text{JSD}(p^{(l)}_{\text{orig}} \| p^{(l)}_{\text{pert}})$ is the divergence between logit lens projections at layer $l$, and $l_0$ is the first divergence layer. CAI $\in [0, 1]$: values near 1 indicate persistent amplification, values near 0 indicate self-correction.

\section{Experimental Setup}

We evaluate three open-weight instruction-tuned LLMs: \textbf{Mistral-7B-Instruct-v0.2} \citep{jiang2023mistral7b} (7B parameters, 32 layers), \textbf{Llama-3-8B-Instruct} \citep{grattafiori2024llama3herdmodels} (8B parameters, 32 layers), and \textbf{Qwen2.5-7B-Instruct} \citep{qwen2025qwen25technicalreport} (7B parameters, 28 layers). All models are loaded in float16 precision with greedy decoding.

For each model, we select the first 60 flipped samples as the mechanistic analysis subset. Logit lens is applied to all 677 pairs; activation patching and component ablation are applied to the 60-sample subset. All experiments are conducted on a university HPC cluster with NVIDIA GPUs.

\section{Results}

\subsection{Perturbation Robustness}

\begin{table}[t]
  \centering
  \begin{tabular}{lccc}
    \toprule
    \textbf{Model} & \textbf{Overall} & \textbf{Name} & \textbf{Number} \\
    \midrule
    Mistral-7B  & 45.1\% & 27.3\% & 63.1\% \\
    Llama-3-8B  & 33.5\% & 23.5\% & 43.8\% \\
    Qwen2.5-7B  & 28.8\% & 18.2\% & 39.6\% \\
    \bottomrule
  \end{tabular}
  \caption{Flip rates by model and perturbation type.}
  \label{tab:flip_rates}
\end{table}

All three models exhibit substantial sensitivity to meaning-preserving perturbations (Table~\ref{tab:flip_rates}). Number format paraphrasing is consistently more disruptive, with the disparity particularly pronounced for Mistral-7B \citep{jiang2023mistral7b} (63.1\% vs.\ 27.3\%).

\subsection{Logit Lens Analysis}

\begin{table}[t]
  \centering
  \begin{tabular}{lcccc}
    \toprule
    \textbf{Model} & \textbf{Flip} & \textbf{No-Flip} & \textbf{$p$} & \textbf{$r$} \\
    \midrule
    Mistral-7B  & 9.1  & 11.6 & 2.07e-04 & 0.157 \\
    Llama-3-8B  & 13.0 & 14.7 & 2.14e-02 & 0.451 \\
    Qwen2.5-7B  & 17.2 & 19.5 & 9.23e-06 & 0.395 \\
    \bottomrule
  \end{tabular}
  \caption{Mean first divergence layer for flipped vs.\ non-flipped samples. All comparisons significant at $p < 0.05$.}
  \label{tab:logit_lens}
\end{table}

Flipped samples diverge at significantly earlier layers across all three models (Table~\ref{tab:logit_lens}). Mistral-7B shows the earliest divergence (layer 9.1), consistent with its highest flip rate.

\begin{figure}[t]
  \centering
  \includegraphics[width=\columnwidth]{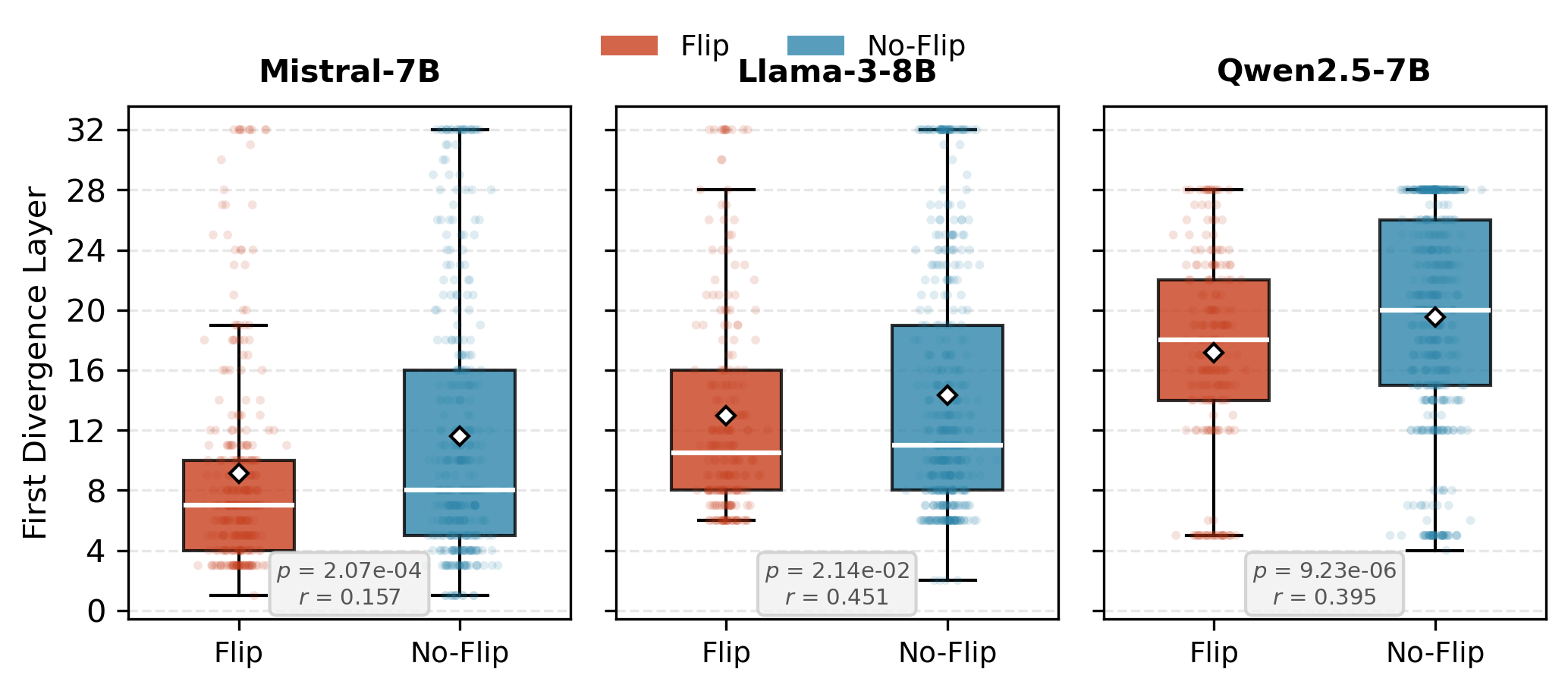}
  \caption{Distribution of first divergence layers for flipped vs.\ non-flipped samples.}
  \label{fig:logit_lens}
\end{figure}

\subsection{Activation Patching}

\begin{table}[t]
  \centering
  \small
  \setlength{\tabcolsep}{5pt}
  \begin{tabular}{lcc}
    \toprule
    Model & Recovery & Rate \\
    \midrule
    Mistral-7B  & 3/60  & 5.0\% \\
    Llama-3-8B  & 43/60 & 71.7\% \\
    Qwen2.5-7B  & 0/60  & 0.0\% \\
    \bottomrule
  \end{tabular}
  \caption{Activation patching recovery rates.}
  \label{tab:patching}
\end{table}

Activation patching reveals a stark architectural divide (Table~\ref{tab:patching}): Llama-3 is highly recoverable (71.7\%), whereas Mistral (5.0\%) and Qwen (0.0\%) are not.

\begin{figure}[t]
  \centering
  \includegraphics[width=0.88\columnwidth]{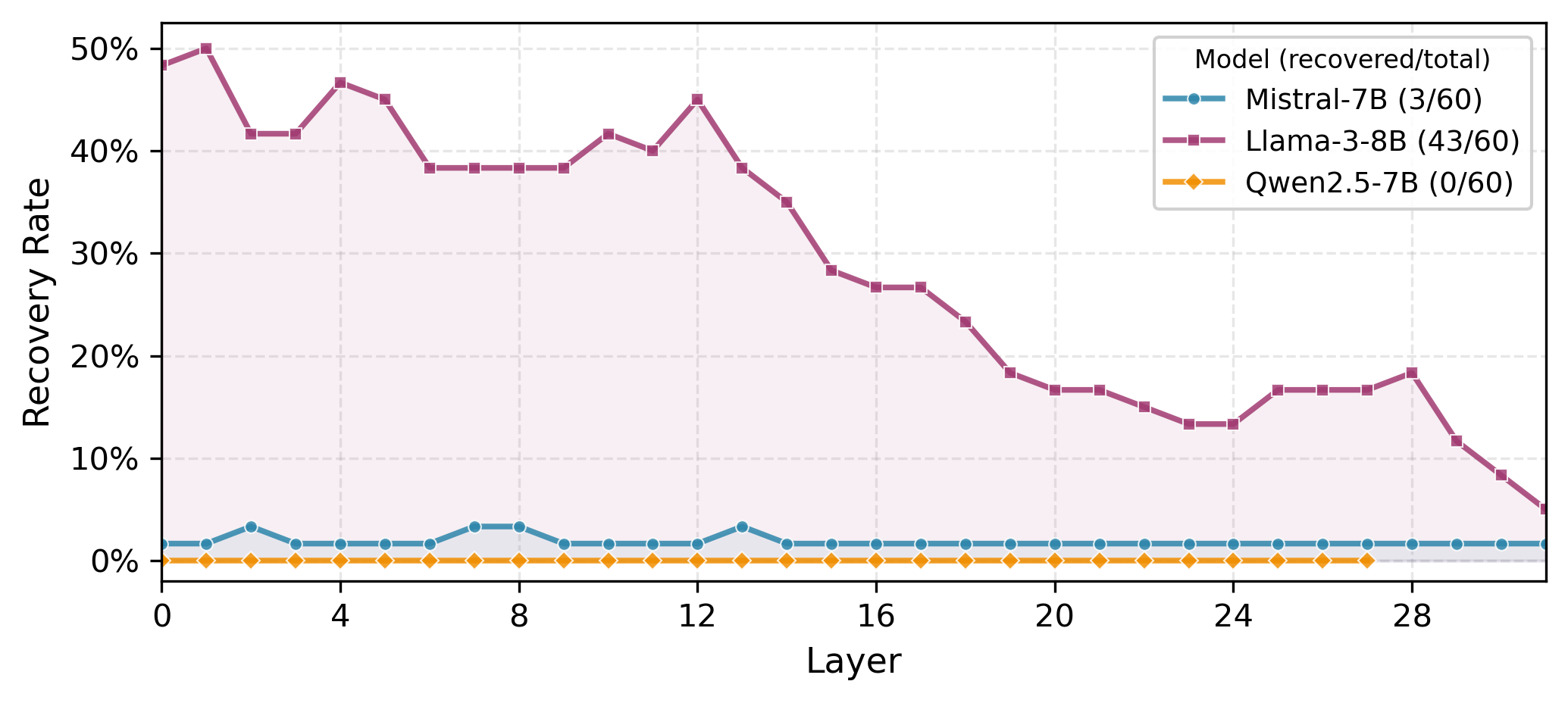}
  \caption{Activation patching recovery rate by layer.}
  \label{fig:patching}
\end{figure}

\subsection{Component Ablation}

\begin{table*}[t]
  \centering
  \small
  \setlength{\tabcolsep}{4pt}
  \begin{tabular}{lcccccc}
    \toprule
    Model & Attn Recov. & MLP Recov. & Attn Mean & MLP Mean & Attn First & MLP First \\
    \midrule
    Mistral-7B & 27/60 (45\%) & 21/60 (35\%) & 1.13 & 0.78 & 11.8 & 14.9 \\
    Llama-3-8B & 45/60 (75\%) & 46/60 (77\%) & 5.78 & 4.95 & 6.4 & 8.0 \\
    Qwen2.5-7B & 24/60 (40\%) & 29/60 (48\%) & 1.75 & 2.15 & 5.2 & 3.9 \\
    \bottomrule
  \end{tabular}
  \caption{Component ablation results. Recov.\ = samples with at least one recovering layer. Mean = mean recovering layers per sample. First = mean first recovery layer.}
  \label{tab:ablation}
\end{table*}

Component ablation reveals architecture-specific patterns (Table~\ref{tab:ablation}). In Mistral, attention is the dominant causal component (27/60 vs.\ 21/60, earlier recovery at layer 11.8 vs.\ 14.9). Llama-3 shows high recovery from both components, consistent with localized failures. Qwen reverses the pattern: MLP sublayers recover more samples (29/60 vs.\ 24/60) at earlier layers (3.9 vs.\ 5.2).

\begin{figure*}[t]
  \centering
  \includegraphics[width=0.88\textwidth]{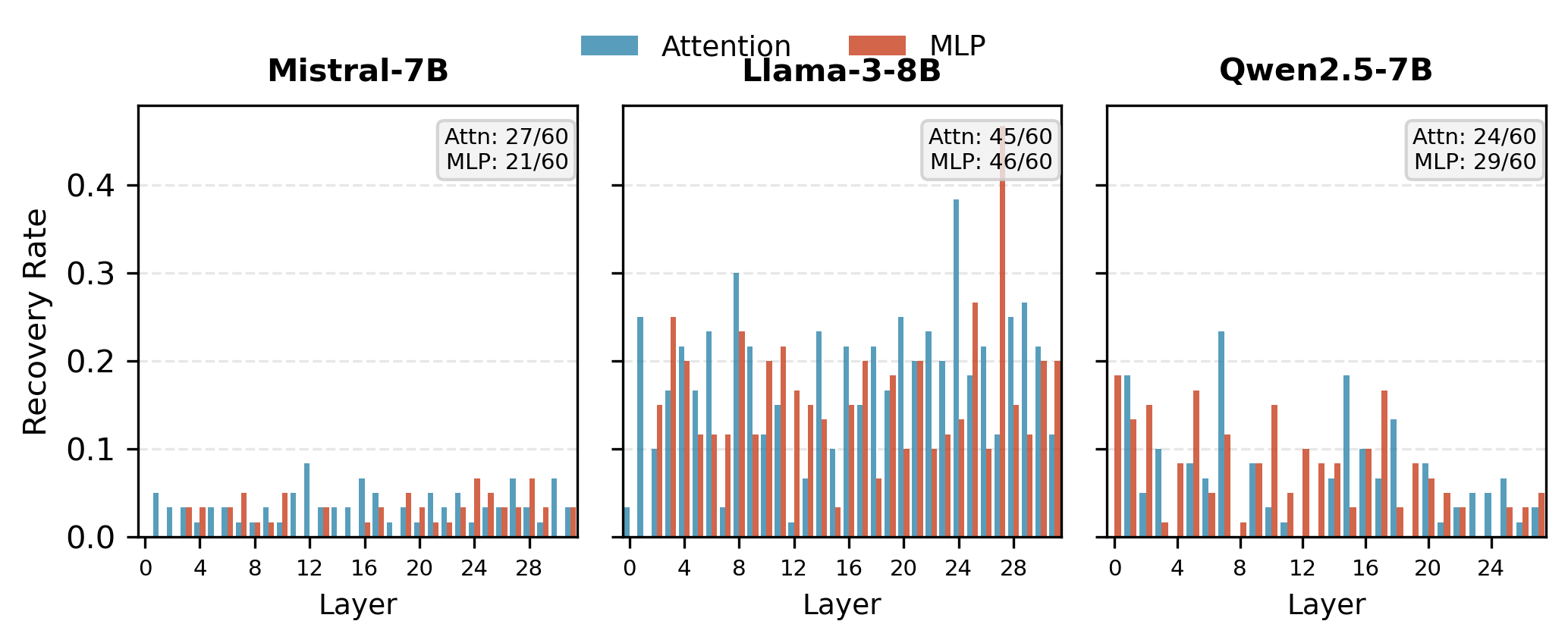}
  \caption{Per-layer recovery under component ablation. Mistral and Llama-3 show attention dominance, while Qwen shows MLP dominance.}
  \label{fig:ablation}
\end{figure*}

\section{Discussion}

\subsection{Cascading Amplification}

\begin{figure}[t]
  \centering
  \includegraphics[width=0.88\columnwidth]{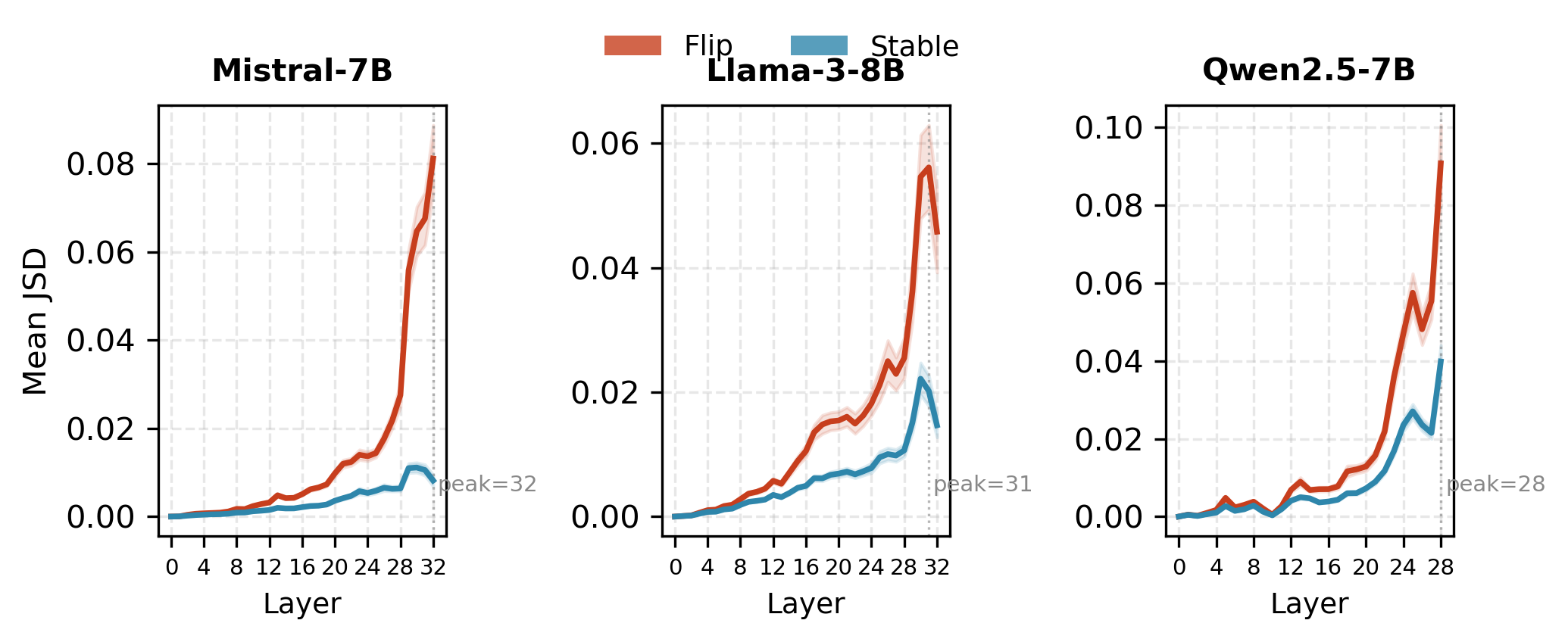}
  \caption{Mean layer-wise JSD trajectory for flipped vs.\ stable samples. Flipped samples exhibit cascading amplification; stable samples show attenuation.}
  \label{fig:jsd_trajectory}
\end{figure}

Figure~\ref{fig:jsd_trajectory} shows the layer-wise JSD trajectory for flipped vs.\ stable samples. For stable samples, JSD remains low as the model attenuates the perturbation. For flipped samples, JSD amplifies through later layers, with the flip-to-stable ratio growing from ${\sim}1.7\times$ at layer 4 to $9.9\times$ at the final layer for Mistral \citep{jiang2023mistral7b}.

Flipped samples exhibit significantly higher CAI across all three models (Table~\ref{tab:cai}). CAI outperforms first divergence layer as a failure predictor for Mistral (AUC 0.679 vs.\ 0.579) and Llama-3 \citep{grattafiori2024llama3herdmodels} (0.593 vs.\ 0.540), while first divergence layer is slightly more predictive for Qwen \citep{qwen2025qwen25technicalreport} (0.605 vs.\ 0.560).

\begin{table}[t]
  \centering
  \small
  \begin{tabular}{lcccc}
    \toprule
    \textbf{Model} & \textbf{Flip} & \textbf{Stable} & \textbf{$p$} & \textbf{$r$} \\
    \midrule
    Mistral-7B  & 0.616 & 0.545 & 2.86e-15 & 0.357 \\
    Llama-3-8B  & 0.548 & 0.511 & 7.39e-05 & 0.186 \\
    Qwen2.5-7B  & 0.500 & 0.472 & 1.06e-02 & 0.121 \\
    \bottomrule
  \end{tabular}
  \caption{CAI for flipped vs.\ stable samples.}
  \label{tab:cai}
\end{table}

This cascading dynamic explains the patching results: in Llama-3, the cascade passes through a narrow bottleneck that patching can interrupt; in Mistral and Qwen, it spreads through redundant pathways.

\subsection{Mechanistic Failure Taxonomy}

Our analysis reveals three qualitatively distinct failure profiles (Table~\ref{tab:failure_profiles}).

\begin{table}[t]
  \centering
  \small
  \begin{tabular}{lccc}
    \toprule
    \textbf{Property} & \textbf{Mistral} & \textbf{Llama-3} & \textbf{Qwen} \\
    \midrule
    Flip rate        & 45.1\% & 33.5\% & 28.8\% \\
    Patch recovery   & 5.0\%  & 71.7\% & 0.0\%  \\
    Dominant comp.   & Attn   & Attn   & MLP    \\
    First diverge    & 9.1    & 13.0   & 17.2   \\
    Failure type     & Distributed & Localized & Entangled \\
    \bottomrule
  \end{tabular}
  \caption{Architecture-specific failure profiles.}
  \label{tab:failure_profiles}
\end{table}

\paragraph{Localized (Llama-3).} Moderate flip rate (33.5\%) but highly recoverable failures (71.7\% patching recovery). Perturbation-sensitive information flows through a narrow computational pathway.

\paragraph{Distributed (Mistral).} Highest flip rate (45.1\%) with near-zero patching recovery (5.0\%). Attention heads at middle-to-late layers are disproportionately causal, but the effect is redundantly encoded across layers.

\paragraph{Entangled (Qwen).} Zero patching recovery with reversed component dominance: MLP sublayers, not attention, are the primary causal components. Perturbation sensitivity is distributed across both layers and components.

We formalize these as: \textbf{Localized} ($R_{\text{patch}} > 0.5$), \textbf{Distributed} ($R_{\text{patch}} < 0.1$, $R_{\text{attn}} > R_{\text{mlp}}$), and \textbf{Entangled} ($R_{\text{patch}} < 0.1$, $R_{\text{mlp}} > R_{\text{attn}}$).

\subsection{Taxonomy Validation via Targeted Repair}
\label{sec:repair}

To validate that the taxonomy is actionable, we implement targeted repair strategies guided by each model's failure profile and evaluate whether repair difficulty follows the predicted ordering: localized $>$ distributed $\approx$ entangled.

\begin{table}[!t]
  \centering
  \small
  \begin{tabular}{llcc}
    \toprule
    \textbf{Model} & \textbf{Strategy} & \textbf{Recovery} & \textbf{Rate} \\
    \midrule
    \multirow{2}{*}{Llama-3}
      & Layer FT (L3--5)        & 25/222  & 11.3\% \\
      & Steering ($\alpha$=0.5) & 27/222  & 12.2\% \\
    \midrule
    \multirow{3}{*}{Qwen}
      & Layer FT (L0,2,5)       & 14/195  & 7.2\%  \\
      & Steering ($\alpha$=1.0) & 14/195  & 7.2\%  \\
      & ROME MLP edit           & $-$2/180 & $-$1.1\% \\
    \midrule
    \multirow{2}{*}{Mistral}
      & Layer FT (L12,14,27)    & 15/305  & 4.9\%  \\
      & Steering ($\alpha$=0.5) & 16/305  & 5.2\%  \\
    \bottomrule
  \end{tabular}
  \caption{Repair results. Recovery = previously-flipped samples now correct. Negative recovery indicates regression.}
  \label{tab:repair_results}
\end{table}

\paragraph{Layer Fine-Tuning.} We freeze all parameters except mechanistically-identified target layers and fine-tune with a combined cross-entropy and representational alignment loss. For Llama-3, fine-tuning layers 3--5 (8.2\% of parameters) recovers 11.3\% of failures.

\paragraph{Steering Vectors.} We compute per-layer correction vectors as the mean hidden-state difference between original and perturbed inputs \citep{turner2024steeringlanguagemodelsactivation}, requiring no training. Steering recovers 12.2\% for Llama-3, 7.2\% for Qwen, and 5.2\% for Mistral.

\paragraph{ROME Editing.} Rank-one MLP weight updates \citep{meng2023locatingeditingfactualassociations} applied to Qwen produce negative recovery ($-$1.1\%), with most edited samples generating malformed outputs, demonstrating that entangled failures resist localized weight surgery.

\paragraph{Results.} The repair results (Table~\ref{tab:repair_results}) confirm the taxonomy's predictive validity. Localized failures (Llama-3) are the most repairable: two independent methods each recover $>$11\%. Entangled failures (Qwen) plateau at $\sim$7\% regardless of method. Distributed failures (Mistral) yield the lowest recovery (5.2\%), consistent with the absence of any single correctable bottleneck.

\section{Conclusion}

We presented a mechanistic analysis of LLM sensitivity to meaning-preserving perturbations. By combining logit lens, activation patching, component ablation, and CAI into the MPD framework, we diagnosed \textit{where} and \textit{how} failures arise within model internals.

Our central finding is that reasoning fragility manifests through architecture-specific failure modes, namely localized (Llama-3 \citep{grattafiori2024llama3herdmodels}, 71.7\% patching recovery), distributed (Mistral \citep{jiang2023mistral7b}, 5.0\%), and entangled (Qwen \citep{qwen2025qwen25technicalreport}, 0.0\%), which we formalize as a mechanistic taxonomy. The Cascading Amplification Index significantly predicts failure across all three architectures (AUC up to 0.679). Critically, the taxonomy predicts repair difficulty: targeted interventions recover 12.2\% of localized failures but only 5--7\% of distributed and entangled failures, validating the taxonomy as actionable.

Future work should explore whether these failure modes persist at larger scales, generalize to other reasoning domains, and whether combining multiple taxonomy-guided interventions can improve recovery rates.

\section*{Limitations}

Our mechanistic analysis is restricted to 60 flipped samples per model due to computational constraints. We evaluate only 7--8B parameter models; scaling behavior is unknown. Our perturbation types are limited to name substitution and number paraphrasing. Component ablation uses zero-ablation, which may not reflect natural causal structure. Finally, our analysis focuses on the answer token position; perturbation effects may also manifest at earlier reasoning steps.

\section*{LLM Usage Disclosure}

We used an LLM assistant to help with experiment implementation and paper drafting. All experimental results were independently verified by the authors, and all scientific claims and interpretations are the authors' own.

\section*{Ethics Statement}
Our work studies existing open-weight models on a public benchmark (GSM8K) and does not involve human subjects. Name substitution pools are designed to avoid demographic bias. All perturbations are meaning-preserving and do not introduce harmful content. We will release our perturbation dataset and analysis code upon acceptance to support reproducibility.
\bibliography{colm2026_conference}
\bibliographystyle{colm2026_conference}

\appendix
\section{Repair Details}

Table~\ref{tab:repair_by_type} breaks down repair recovery by perturbation type.

\begin{table}[H]
  \centering
  \small
  \begin{tabular}{llcc}
    \toprule
    \textbf{Model} & \textbf{Pert.\ Type} & \textbf{FT} & \textbf{Steering} \\
    \midrule
    \multirow{2}{*}{Llama-3}
      & Name swap   & 18.2\% & 13.0\% \\
      & Num.\ para. & 7.6\%  & 11.7\% \\
    \midrule
    \multirow{2}{*}{Qwen}
      & Name swap   & 8.1\%  & 8.1\%  \\
      & Num.\ para. & 6.8\%  & 6.8\%  \\
    \midrule
    \multirow{2}{*}{Mistral}
      & Name swap   & 6.5\%  & 6.5\%  \\
      & Num.\ para. & 4.2\%  & 4.7\%  \\
    \bottomrule
  \end{tabular}
  \caption{Repair recovery rate by perturbation type.}
  \label{tab:repair_by_type}
\end{table}

Table~\ref{tab:ablation_by_type} breaks down component ablation by perturbation type.

\begin{table}[H]
  \centering
  \small
  \begin{tabular}{llcc}
    \toprule
    \textbf{Model} & \textbf{Pert.\ Type} & \textbf{Attn} & \textbf{MLP} \\
    \midrule
    \multirow{2}{*}{Mistral-7B}
      & Name swap   & 0.82 & 0.41 \\
      & Num.\ para. & 1.26 & 0.93 \\
    \midrule
    \multirow{2}{*}{Llama-3-8B}
      & Name swap   & 7.00 & 6.17 \\
      & Num.\ para. & 5.03 & 4.19 \\
    \midrule
    \multirow{2}{*}{Qwen2.5-7B}
      & Name swap   & 0.93 & 2.33 \\
      & Num.\ para. & 2.02 & 2.09 \\
    \bottomrule
  \end{tabular}
  \caption{Mean recovering layers per sample by perturbation type and component.}
  \label{tab:ablation_by_type}
\end{table}

\section{Reproducibility}
All experiments use HuggingFace Transformers with \texttt{device\_map="auto"} and float16 precision. Greedy decoding (temperature = 0) ensures deterministic outputs. Logit lens, activation patching, and component ablation are implemented via forward hooks without modifying model weights. CAI is computed from logit lens outputs with no additional model inference required. Perturbation dataset construction uses rule-based transformations with manual validation. Statistical tests use SciPy's Mann-Whitney $U$ implementation. Repair experiments use AdamW (lr = 1e-5, 3 epochs) for layer fine-tuning and mean activation differences over 100 calibration samples for steering vectors. Code and data will be released upon acceptance.
\end{document}